\documentclass[10pt,twocolumn,letterpaper]{article}

\usepackage{cvpr}              

\usepackage[dvipsnames]{xcolor}

\usepackage[utf8]{inputenc} 
\usepackage[T1]{fontenc}    
\usepackage{url}            
\usepackage{booktabs}       
\usepackage{amsfonts}       
\usepackage{nicefrac}       
\usepackage{microtype}      
\usepackage{xcolor}         

\usepackage{float}
\usepackage{wrapfig}

\usepackage{multirow}
\usepackage{graphicx}
\usepackage{amsmath}
\usepackage{colortbl}

\renewcommand\footnotemark{}

\definecolor{cvprblue}{rgb}{0.21,0.49,0.74}
\usepackage[pagebackref,breaklinks,colorlinks,citecolor=cvprblue]{hyperref}

\def\paperID{2932} 
\def\confName{CVPR}
\def\confYear{2024}

\title{Action Detection via an Image Diffusion Process}

\author{Lin Geng Foo\textsuperscript{1}\thanks{ \dag~Corresponding author}
~~~ Tianjiao Li\textsuperscript{1}
~~~ Hossein Rahmani\textsuperscript{2} ~~~ Jun Liu\textsuperscript{1\dag} \\
\textsuperscript{1}Singapore University of Technology and Design ~~ 
\textsuperscript{2}Lancaster University\\
{\tt\small \{lingeng\_foo,tianjiao\_li\}@mymail.sutd.edu.sg,}\\ 
{\tt\small h.rahmani@lancaster.ac.uk, jun\_liu@sutd.edu.sg } \\
}

\begin{document}
\maketitle

\begin{abstract}
Action detection aims to localize the starting and ending points of action instances in untrimmed videos, and predict the classes of those instances. In this paper, we make the observation that the outputs of the action detection task can be formulated as images. Thus, from a novel perspective, we tackle action detection via a three-image generation process to generate starting point, ending point and action-class predictions as images via our proposed Action Detection Image Diffusion (ADI-Diff) framework. Furthermore, since our images differ from natural images and exhibit special properties, we further explore a Discrete Action-Detection Diffusion Process and a Row-Column Transformer design to better handle their processing. Our ADI-Diff framework achieves state-of-the-art results on two widely-used datasets.
\end{abstract}

\vspace{-3.5mm}
\section{Introduction}

The goal of action detection is to localize the starting and ending points of action instances in untrimmed videos, while also predicting the classes of those actions.
Action detection is important across many video analysis applications, including healthcare monitoring \citep{sathyanarayana2018vision,nweke2019data}, sports analysis \citep{jiang2016automatic,giancola2018soccernet} and security surveillance \citep{vahdani2022deep,amrutha2020deep}, and has attracted a lot of research attention.
A common approach \citep{xu2017r,gao2017turn,chen2022dcan,lin2021learning,weng2022efficient} is to first extract proposals of action instances, before processing each of these proposals individually to produce refined starting point, ending point, and action-class predictions.
Many works focus on improving the action proposal localization process (e.g., with a better starting and ending point regression head \cite{lin2018bsn,lin2019bmn,liu2019multi}), or designing better model architectures (e.g., graph models \cite{xu2020g,li2020graph,zhao2021video} and Transformers \cite{zhang2022actionformer,liu2022end}).
Nevertheless, action detection still remains challenging, since actions often contain complex motions with high intra-class variability \cite{zhu2021enriching,guo2022uncertainty}, and difficulties also arise due to the varying lighting conditions, different viewpoints and background clutter \citep{guo2022uncertainty,lin2021learning,shaikh2021rgb}.

On the other hand, image diffusion models \cite{ho2020denoising,song2020improved} have recently undergone rapid development and show an excellent capability to generate high-quality images.
Image diffusion models aim to obtain a high-quality image from an noisy and uncertain image, and achieve this via step-by-step progressive denoising.
Intuitively, the diffusion model's process of progressive denoising helps to bridge the large gap between the high-quality target images and the noisy images by breaking it down into smaller intermediate steps \citep{song2019generative}, which assists the model in converging towards generating the high-quality target images.
Thus, image diffusion models \cite{ho2020denoising,song2020improved} can improve image quality and training stability, and possess a strong ability to generate high-quality target images that align well with the provided input conditions.

\begin{figure}[t]
    \centering  
     \includegraphics[width=\linewidth]{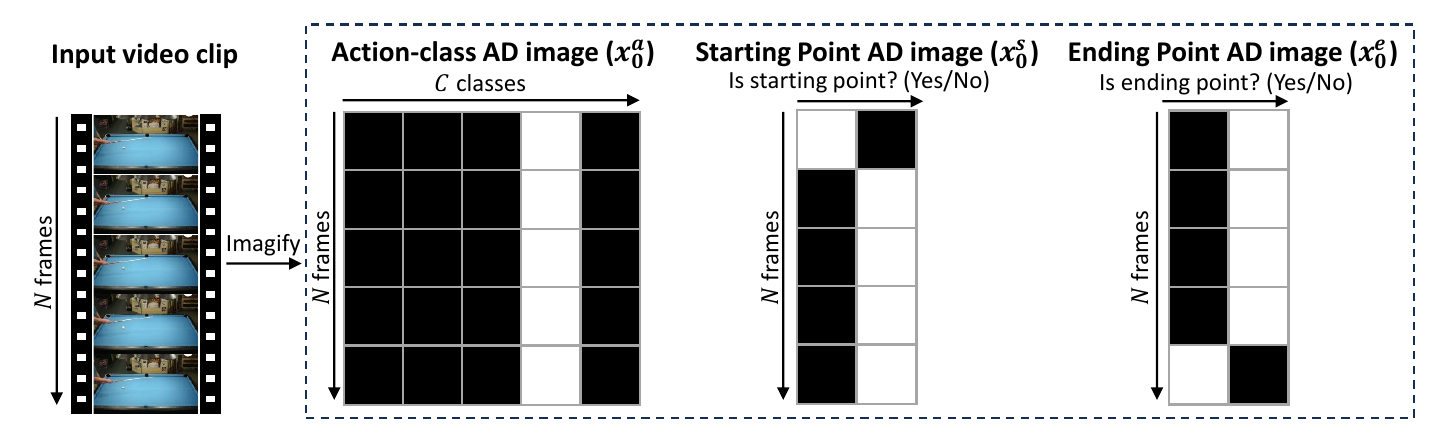}           
    \vspace{-7mm}    
    \caption{
    Illustration of our formulated AD images, which allow us to tackle action detection by generating three images.    
    The action-class AD image ($x^a$) has a shape of $N \times C$, while the starting and ending point AD images ($x^s$ and $x^e$) both have a shape of $N \times 2$, where we show $N =5$ and $C=5$ in this figure for illustration.       
    Specifically, the pixel values in a row of the image form the probabilities of a discrete distribution regarding a specific video frame, e.g., the $n$-th row of the action-class AD image represents the probability distribution over the action classes for the $n$-th frame.    
     We depict the ground truth AD images ($x^a_0,x^s_0,x^e_0$) in this figure, thus each row contains a single white pixel (with value 1) in each row depicting the correct prediction, while the other pixels are black in color (with value 0).
    }
    \label{fig:image_formulation}
    \vspace{-7mm}
\end{figure}

In this work, inspired by the efficacy of image diffusion models, we make the following observation: the three outputs (starting point, ending point and action-class) for the action detection task can be \textit{formulated as images}.
For instance, the action-class predictions can be represented by a $N \times C$ image (where $N$ is the number of frames and $C$ is the number of action-classes), while the starting and ending point predictions can each be represented by a $N \times 2$ image, as shown in Fig.~\ref{fig:image_formulation}.
Hence, from a new perspective, we can re-cast action detection as a three-image generation task, tackled by generating these three ``action detection'' images -- which we call \textit{AD images} -- as output.
Then, in order to generate these AD images with a high level of quality, we can leverage image diffusion models \citep{sohl2015deep,ho2020denoising,song2021denoising} with their strong image generation capabilities.

\begin{figure*}
    \centering  
     \includegraphics[width=\linewidth]{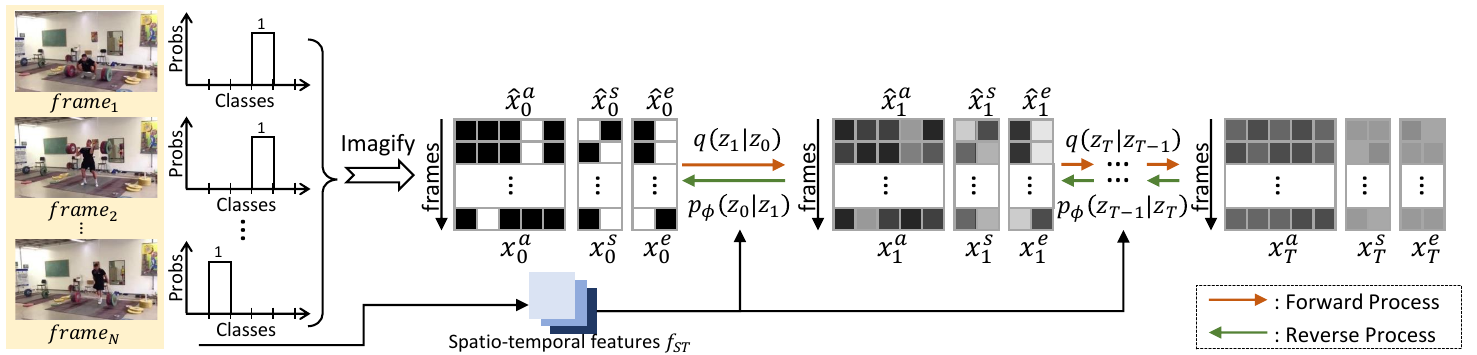}           
    \vspace{-7mm}    
    \caption{
    Illustration of the proposed AD Image Diffusion (ADI-Diff) framework. 
    The forward process (represented with orange arrows) progressively diffuses the ground truth AD images $x_0^a,x_0^s,x_0^e$ towards a noisy outcome, which generates supervisory signals for intermediate steps.
    On the other hand, the reverse process (represented with green arrows) is trained to denoise the noisy inputs $x_T^a,x_T^s,x_T^e$ while conditioned on extracted spatio-temporal features $f_{ST}$ from the input video, to obtain the output AD images $\hat{x}_0^a,\hat{x}_0^s,\hat{x}_0^e$.
    }
    \label{fig:overall_network}
    \vspace{-5mm}
\end{figure*}

To this end, we propose an AD Image Diffusion (ADI-Diff) framework for action detection, as shown in Fig.~\ref{fig:overall_network}. 
Our ADI-Diff framework learns to generate the target high-quality action-class, starting point and ending point AD images via diffusion. 
Following previous works on diffusion models \citep{ho2020denoising,song2021denoising}, our ADI-Diff framework comprises two opposite diffusion processes: the \textit{forward process} and the \textit{reverse process}.
Specifically, the forward process aims to generate supervisory signals of intermediate steps during training, through progressively adding noise to the ground truth AD images.
Conversely, the reverse process aims to learn to reverse the forward process, i.e, by learning to denoise and produce high-quality AD images, which is the main part of our action detection pipeline.

However, directly using standard diffusion models \citep{ho2020denoising,song2021denoising} for our ADI-Diff framework can be sub-optimal since they learn to generate natural images, while our proposed AD images differ from natural images, because AD images also represent a set of \textit{discrete probability distributions}.
For instance, our AD images are used to tackle a classification problem (either among $C$ action-classes or 2 classes for starting/ending point predictions), which is a problem of predicting discrete probability distributions.
Thus, our AD images also represent a set of discrete probability distributions, where the pixels in each row of the image represent the probabilities of a discrete distribution.
Hence, instead of following the standard diffusion process to map between a high-quality image and a totally uncertain image, our diffusion process should learn to map between the ground truth -- which is an \textit{ideal discrete probability distribution} -- and a \textit{totally uncertain discrete probability distribution}.
In other words, the standard diffusion process, which progressively introduces Gaussian noise in the forward process and converges towards Gaussian noise, is not well-suited for our needs.
Therefore, we propose a novel Discrete Action-Detection Diffusion Process that constrains each forward diffusion step to produce discrete probability distributions, which enables us to generate the desired high-quality AD images from the noisy and uncertain probability distributions more effectively.

Moreover, in contrast to traditional images which contain rich local spatial correlations in both dimensions, our AD images exhibit different relationship patterns along each of the two dimensions.
Specifically, in our AD images, there is a strong sequential ordering between adjacent rows (i.e., between temporal frames), which differs from the inter-class relationships between adjacent columns (e.g., between action classes).
Hence, since our AD images differ from traditional images, existing diffusion network designs, which tend to focus on 2D spatial processing in local neighbourhoods, are not suitable for our use. 
Thus, we further propose our Row-Column Transformer design for our diffusion model to effectively extract class information across the columns while encoding temporal relationships across the rows.

In summary, our contributions are as follows: 
\textbf{(1)} From a novel perspective, we re-cast action detection as a three-image generation problem and generate the AD image predictions via our AD Image Diffusion (ADI-Diff) framework. 
\textbf{(2)} We propose a Discrete Action-Detection Diffusion Process that constrains the forward diffusion process to produce discrete probability distributions, which provides a good mapping between the input noisy distribution and the ground truth distribution. 
\textbf{(3)} To handle our AD images which are different from traditional images, we further introduce a Row-Column Transformer design for our diffusion network.

\section{Related Work}

In \textbf{action detection}, many approaches \citep{xu2017r,gao2017turn,chen2022dcan,lin2021learning,weng2022efficient} first extract action proposals before processing them individually to predict action classes and refined starting and ending points. 
These methods are generally split into two categories: anchor-based and anchor-free.
Anchor-based methods \citep{xu2017r,gao2017turn,chen2022dcan} such as multi-tower networks \citep{chao2018rethinking} and temporal feature pyramid networks \citep{liu2020progressive,liu2019multi} generate a dense set of anchors with pre-defined lengths throughout the video that act as action proposals.
On the other hand, anchor-free methods \citep{lin2021learning,weng2022efficient,nag2022proposal,zhang2022actionformer,nag2023difftad} often predict actionness scores \citep{zhao2017temporal,wang2016actionness} or action boundary confidence scores \citep{lin2018bsn,lin2021learning}
for video frames in order to generate action proposals.
Besides, some existing works explore different architectures to encode spatio-temporal information, including RNNs \citep{buch2017sst,yeung2016end}, graph models \citep{zhao2021video,zeng2019graph,xu2020g,li2020graph,bai2020boundary}, or Transformers \citep{tan2021relaxed,nawhal2021activity,chang2021augmented,liu2022end}.
Moreover, some proposal-free methods \cite{nag2022proposal,nag2022semi} have also been explored recently, and our ADI-Diff also falls into this category.
Different from previous works, we re-cast action detection as a three-image generation problem, and leverage the diffusion model's strong image generation capability to generate the three AD image predictions.
Our proposed ADI-Diff framework effectively handles the challenging action detection task by generating high-quality starting/ending point and action-class AD images, achieving good results.

\textbf{Diffusion models} have emerged as an effective way to sample from a data distribution by learning to estimate the gradients of the data distribution \citep{song2019generative}. 
Originally introduced in the context of image generation \citep{sohl2015deep}, diffusion models have seen much development in recent years \citep{ho2020denoising,song2021denoising,austin2021structured,foo2023ai,xu20236d}, and have been explored across various generation tasks, including video \citep{singer2022make}, point cloud \citep{luo2021diffusion} and text \citep{li2022diffusion} generation.
Diffusion models have also been adopted for human activity analysis  \cite{gong2023diffpose,shan2023diffusion,liu2023diffusion,foo2023distribution}, 
e.g., for pose estimation \cite{gong2023diffpose} where a GMM-based forward process is proposed.
Moreover, some works \cite{chen2023diffusiondet} adopt diffusion models for regressing the bounding box of an object, which has also inspired a similar process for regressing temporal boundaries \cite{nag2023difftad}.
In this work, considering that action detection requires starting point, ending point and action-class outputs which can be treated as three AD images, and that diffusion models naturally have a strong image generation capability, we propose to cast action detection as a three-image diffusion process.
At the same time, since AD images are not natural images and have their own properties, we propose modifications to the image diffusion process, which attain good performance.

\section{ADI-Diff Framework}

In this paper, we tackle the action detection task by \textit{casting it as an image generation problem} (as described in Sec.~\ref{sec:action_detection_formulation}), and leverage a diffusion model via our proposed ADI-Diff framework to generate AD images that encode the required starting point, ending point and action-class information.
Moreover, in Sec.~\ref{sec:discrete_diffusion}, we design a Discrete Action-Detection Diffusion Process, constraining the pixels along each row of our AD images to form a discrete probability distribution in the forward process.
Besides, in order to perform diffusion for our AD images which exhibit different relationship patterns along each of the two dimensions (i.e., between frames and between classes), we propose a Row-Column Transformer architecture in Sec.~\ref{sec:transformer_architecture}.

\subsection{Formulation of AD images}
\label{sec:action_detection_formulation}

In this work, we observe that we can reformulate the three outputs of the action detection task (starting point, ending point and action-class prediction) as three images.
Hence, from a new perspective, we can cast action detection as an image diffusion process to generate these three AD images.
Below, we describe how we formulate our action-class, starting point and ending point AD images to encode the corresponding predictions.

\textbf{Action-class AD image $x^a$.}
As shown in Fig.~\ref{fig:image_formulation}, to capture the action-class predictions at each time step, we set the action-class AD image to be a matrix $x^a$ with shape $N \times C$, where $N$ is the number of frames in the video and $C$ represents the number of action classes.
The matrix $x^a$ can be seen as an image, where the pixel value at the $n$-th row and $c$-th column is the probability that action $c$ is happening at the $n$-th frame of the video, and is constrained to be in $[0,1]$.
Thus, $x^a$ is a grayscale image with shape $N \times C$, which can be generated via an image diffusion process.

\textbf{Starting Point and Ending Point AD images $x^s,x^e$.}
Next, in order to encode temporal boundary predictions, we produce two AD images: the starting point AD image $x^s \in [0,1]^{N \times 2}$ and the ending point AD image $x^e \in [0,1]^{N \times 2}$, which respectively encode predictions of the starting points and ending points of action instances.
Specifically, the $N$ rows of the matrix $x^s$ (or $x^e$) encode information regarding the $N$ frames, with the $2$ pixels in each row respectively encoding the probabilities for the presence and absence of a starting (or ending) point.
For example, for the starting point AD image $x^s$, the pixel value in the first column of the $n$-th row represents the probability of a starting point occurring at the $n$-th frame.
The same goes for the ending point AD image $x^e$, except that the pixel value in the first column encodes the probability of an ending point occurring instead. 
In summary, both $x^s$ and $x^e$ can be seen as grayscale images with shape $N \times 2$, as shown in Fig.~\ref{fig:image_formulation}.

\subsection{Discrete Action-Detection Diffusion Process}
\label{sec:discrete_diffusion}

After re-casting action detection as a three-image generation task, we seek to generate high-quality AD images to handle action detection effectively.
To achieve this, we derive inspiration from the strong image generation capabilities of diffusion models \cite{ho2020denoising,song2021denoising}, and adopt a diffusion-based approach to generate the three AD images for action detection.
Overall, diffusion models \cite{ho2020denoising,song2021denoising} aim to \textit{obtain a high-quality image from a totally noisy and uncertain image}, and do so by progressively removing the noise and uncertainty over multiple steps.
Specifically, to learn a mapping between a noisy image (that is totally random and uncertain) and a high-quality image, standard diffusion models consist of a \textit{forward process} where Gaussian noise is progressively added to high-quality images.
Meanwhile, the \textit{reverse process} learns to reverse the forward process, i.e., to denoise the noisy and uncertain inputs to obtain high-quality images.
These processes enable diffusion models to bridge the large gap between the input noisy images and the target high-quality images, and \textit{obtain high-quality images from noisy and uncertain images}.

Nevertheless, using standard image-based diffusion models \citep{ho2020denoising,song2021denoising} directly can be sub-optimal.
Specifically, these diffusion models aim to generate natural images from noisy images, so they add Gaussian noise during the forward process to naturally obtain intermediate noisy images. 
However, our AD images differ from natural images, since we are using our AD images to deal with a classification problem, e.g., the action-class AD images are used to tackle a $C$-way action classification task, while the starting point AD images are used to predict starting points as a binary classification task (yes/no), and the same goes for ending point AD images.
In other words, our AD images are in fact a set of \textit{discrete probability distributions}.
Thus, using the Gaussian noise is not suitable, because we want our diffusion process to learn to map between the ground truth -- an ideal \textit{discrete probability distribution} -- and a totally uncertain \textit{discrete probability distribution}, with intermediate discrete distributions to bridge the gap.
In order to form intermediate discrete probability distributions, we cannot simply apply Gaussian noise during the forward process, instead we need to constrain each step of the forward process to produce a \textit{discrete probability distribution}, and  also converge towards a totally uncertain \textit{discrete probability distribution}.
Hence, below we design our own Discrete Action-Detection Diffusion Process.

Firstly, following the standard diffusion model that adds random noise during their forward process to obtain a totally noisy and uncertain image, we would like to add random noise to obtain a totally noisy and uncertain discrete distribution in our forward diffusion process.
We note that, when the classification predictions are totally uncertain, there should be an equal probability of predicting any class, which corresponds to the \textit{Uniform distribution}.
Since the Uniform distribution is the most uncertain, we would like to add noise to converge towards the Uniform distribution in our forward process (\textbf{Property 1}).

Apart from fulfilling Property 1 above, we find that previous diffusion models \cite{ho2020denoising} also have two other important properties that help to facilitate the training of the step-by-step diffusion process. 
Therefore, to maintain the efficacy of the diffusion framework, we here also need to satisfy these two properties, which are as follows:
\textbf{Property 2)} Following previous diffusion models \cite{ho2020denoising}, there should be a formula to efficiently jump over $t$ forward steps, i.e., a formulation for $q(z_t | z_0)$ (as in Eq.~\ref{eq:zt from z0_2}). 
This formula facilitates training and allows us to randomly sample multiple time steps, without having to iterate through many forward steps to get to a specified step.
\textbf{Property 3)} Following previous diffusion models \cite{ho2020denoising}, we also need a formulation for the forward process posterior, i.e., $q(z_{t-1} | z_t, z_0)$ in Eq.~\ref{eq:forward_posterior}, which allows us to generate $z_{t-1}$ from $z_t$ to form a $z_{t-1},z_t$ pair, enabling a direct step-wise comparison against the reverse process step during training \citep{ho2020denoising}.

Therefore, in order to fulfill these properties, and learn to generate the three AD images via our diffusion, we design our forward and reverse process as described below.

\textbf{Forward Process.}
In the forward diffusion process, we aim to create the supervisory signals of intermediate steps for training.
Notably, the forward process of previous works add Gaussian noise at every forward step, which eventually 
 corrupts a natural image (at step $0$) into Gaussian noise (at step $T$).
Here, we instead wish to add a specific type of noise to fulfill Property 1.
Thus, we initialize the ground truth AD images, and then progressively diffuse (the rows of) the ground truth AD images towards the Uniform distribution over $T$ steps.

Next, we formally introduce some definitions.
For simplicity, we describe the diffusion process for a single row of the action-class AD image as an example, which is a discrete probability distribution with $C$ classes.
Specifically, we define $z_t$ to be a discrete probability distribution at step $t$ of the diffusion process, where $z_t$ is a vector of length $C$.
At step $0$ of the diffusion process, $z_0$ represents the ground truth, and is a one-hot vector with value 1 at the index of the ground truth category, and 0 elsewhere.
The forward process spans $T$ steps, where noise is gradually added to $z_0$, such that after $T$ steps, $z_T$ is approximately a Uniform distribution.

Concretely, to create intermediate distributions $\{z_1, ..., z_T \}$ from $z_0$, 
we add random noise $v_t$ at each $t$-th forward step, which progressively makes the prediction more uncertain.
This addition of random noises ($\{ v_t \}_{t=1}^T$) is an important component \citep{song2019generative,ho2020denoising,song2020improved} that facilitates exploration of the low-density regions of the data distribution.
Here, we add $v_t$ at each step according to a \textit{Multinomial distribution} with uniform probability parameters, which crucially allows us to \textit{converge towards the Uniform distribution} and satisfy Property 1 (as explained in further detail later).

Specifically, at each step $t$, to obtain $z_t$ from $z_{t-1}$, we increase the uncertainty by introducing a small chance to randomly select any class (which might not be the correct ground truth class) with equal probability.
Here, we introduce $\beta_t$ as a small positive hyperparameter to control the small increase in randomness at step $t$.
Following the above intuition, we can formulate each $t$-th forward step as:

\vspace{-3mm}
\setlength{\belowdisplayskip}{0pt}
\begin{equation}
    \label{eq:single forward}    
    z_t  =  (1 - \beta_t) z_{t-1} + \beta_t v_t,        
\end{equation}
where $v_t$ is a random vector of length $C$ whose elements are non-negative and add up to 1, i.e., forming the probabilities of a discrete distribution.
We obtain $v_t$ via sampling from a $MN_K (K, \frac{1}{C} \boldsymbol{1} )$ distribution, where $MN$ stands for the Multinomial distribution, $K$ is a hyperparameter for the number of trials, $\frac{1}{C} \boldsymbol{1}$ gives a uniform probability of selecting each class in each trial (where $\boldsymbol{1}$ is a vector of 1's with length $C$),
and we further divide the resulting sample by $K$ to let elements of $v_t$ sum to 1 (denoted by the subscript $K$).

Therefore, the likelihood $q(z_t | z_{t-1})$ of observing $z_t$ given $z_{t-1}$ can be formulated as:

\vspace{-4mm}
\begin{equation}
    \label{eq:single forward_2}  
    q (z_t | z_{t-1}) =  MN_{ \frac{K(z_t - (1- \beta_t) z_{t-1})}{\beta_t}} (z_t; K, \frac{1}{C} \boldsymbol{1}, z_{t-1} ),         
\end{equation}
where, with slight abuse of notation, $MN (z_t;\cdot)$ is the Multinomial's likelihood of observing $z_t$, and the subscript is the substitution formula ($Kv_t$ in terms of $z_t$) which is used to formulate the exact likelihood, such that $Kv_t$ follows the Multinomial distribution (more details in Supp).

Next, expanding upon Eq.~\ref{eq:single forward} which represents a single forward step, the formula for $t$ steps of the forward process starting from $z_0$ can be derived as:
\setlength{\abovedisplayskip}{4pt}
\vspace{-2mm}
\begin{equation}
    \label{eq:zt from z0} 
    z_t  =  \bar{\alpha}_t z_0 +   (\prod_{\tau=2}^t \alpha_\tau) \beta_1 v_1 + (\prod_{\tau=3}^t \alpha_\tau) \beta_2 v_2 + .... + \beta_t v_t,    
\end{equation}
where $\alpha_t = 1 - \beta_t$ and $\bar{\alpha}_t = \prod_{\tau=1}^t \alpha_\tau$.
Then, the corresponding likelihood $q(z_t | z_0)$ of observing $z_t$ (after $t$ steps) can be approximately formulated as the following (with the proof and analysis in Supp):

\vspace{-4mm}
\begin{equation}
    \label{eq:zt from z0_2} 
    q (z_t | z_0) =  MN_{\frac{B_t K(z_t - \bar{\alpha}_t z_0)}{1 - \bar{\alpha}_t}} ( z_t ; B_t K, \frac{1}{C}  \boldsymbol{1}, z_0 ),         
\end{equation}
where $B_t = \frac{(1 - \bar{\alpha}_t)^2}{( (\prod_{\tau=2}^t \alpha_\tau)^2 \beta_1^2    + (\prod_{\tau=3}^t \alpha_\tau)^2 \beta_2^2  + .... + \beta_t^2  )}  $.

Note that, we can show that our diffusion process \textit{fulfills Property 1}.
Specifically, when we set our $\beta_t$'s to be relatively high such that $\bar{\alpha}_t$ converges to 0 as  $t \rightarrow T$, the likelihood in Eq.~\ref{eq:zt from z0_2} converges towards $ q (z_T | z_0) =   MN_{B_T K} (B_T K, \frac{1}{C} \boldsymbol{1} )$. 
Then, using the properties of the Multinomial distribution, we observe that $z_T$ approximately converges to a Uniform distribution in expectation, i.e., $\mathbb{E} [z_T] = \frac{1}{C} \boldsymbol{1}$.
This explicitly shows that our diffusion framework satisfies Property 1.
At the same time, we also \textit{fulfill Property 2}, since Eq.~\ref{eq:zt from z0_2} enables us to directly generate the intermediate distributions $\{z_1, ..., z_T \}$ from $z_0$ for efficient training.

Next, we would like to fulfill Property 3, which will provide a way to obtain $z_{t-1}$ from $z_t$ in the forward process, giving us a pair of $z_{t-1},z_t$ to facilitate the step-wise training of the reverse process \citep{ho2020denoising,song2021denoising,austin2021structured}.
Specifically, this requires a formulation of the forward process posterior $q(z_{t-1} | z_t, z_0)$.
We can formulate a tractable expression for $q(z_{t-1} |z_t, z_0)$ by using the properties of the Markov chain, as follows:

\vspace{-2mm}
\begin{footnotesize}
\begin{align}
    q(z_{t-1} & |z_t, z_0) = \frac{1}{\sigma_t}  \big(   MN_{ \frac{K(z_t - (1- \beta_t) z_{t-1})}{\beta_t}} (z_{t-1}; K, \frac{1}{C} \boldsymbol{1}, z_{t} )    \big)    \nonumber    \\
    \label{eq:forward_posterior}
    &\cdot \big( MN_{\frac{B_{t-1} K ( z_{t-1} - \bar{\alpha}_{t-1} z_0)}{1 - \bar{\alpha}_{t-1}}} ( z_{t-1}; B_{t-1} K, \frac{1}{C}  \boldsymbol{1},z_0 ) \big),       
\end{align}
\end{footnotesize}
where
{\footnotesize $\sigma_t = \sum_{z_{t-1}} \big[ \big(   MN_{ \frac{K(z_t - (1- \beta_t) z_{t-1})}{\beta_t}} (z_{t-1}; K, \frac{1}{C} \boldsymbol{1}, z_{t} )    \big) \cdot \big( MN_{\frac{B_{t-1} K ( z_{t-1} - \bar{\alpha}_{t-1} z_0)}{1 - \bar{\alpha}_{t-1}}} ( z_{t-1}; B_{t-1} K, \frac{1}{C}  \boldsymbol{1},z_0 ) \big) \big]$} (more details in Supp).
Note that, in practice we can fix $\sigma_t$ as a hyperparameter, since it is constant for all observed $z_{t-1}$.

\textbf{Reverse Process.}
Using the forward process presented above, we can generate the intermediate distributions $\{z_1, ..., z_T \}$. 
Then, we can use these intermediate distributions to optimize our diffusion model $d$ (parameterized by $\phi$) to learn the reverse diffusion process.
As shown in Fig.~\ref{fig:overall_network}, the reverse process aims to generate the AD image outputs after $T$ diffusion steps.

First, we extract information from the input video to facilitate the diffusion process.
To do this, we follow previous works \citep{shi2023tridet,zhang2022actionformer,shi2022react,vahdani2022deep,nag2022proposal,lin2021learning} and extract features from video snippets with a pre-trained feature extractor.
Specifically, we extract a feature $f_{ST} \in \mathbb{R}^{N \times C_{ST}}$, where $N$ is the number of frames and $C_{ST}$ is the number of channels.
Each reverse step will be conditioned on this extracted feature $f_{ST}$.

Next, we perform the reverse diffusion process.
We first initialize the input noisy distribution $z_T$ that is approximately a Uniform distribution -- more precisely, $z_T$ follows a $MN_{B_T K} (B_T K, \frac{1}{C} \boldsymbol{1} )$ distribution to match with the $z_T$ of our forward process.
Then, we perform the reverse diffusion process $z_T \rightarrow \hat{z}_{T-1} \rightarrow ... \rightarrow \hat{z}_0$ to generate the predictions $\hat{z}_0$, where the hat $(\;\hat{}\;)$ operator denotes that these are estimates produced by our diffusion model $d$ (and not the forward process).
Specifically, we define our reverse process in a step-by-step manner as follows:

\vspace{-6mm}
\begin{align}
    \label{eq: sample}  
    \hat{z}_{t-1} = d_{\phi} (\hat{z}_t, f_{ST}, f_t),\;\;\; t \in \{1,...,T\},         
\end{align}
where $f_t$ is the unique step embedding to represent the $t^{th}$ diffusion step, which we generate via the sinusoidal function.
By performing the $T$ steps of the reverse process via Eq.~\ref{eq: sample}, we can produce output predictions $\hat{z}_0$ from the uncertain distribution $z_T$.
Moreover, to better represent the intermediate and target distributions, we initialize $M$ samples during training and perform the reverse process on $M$ samples, instead of using a single sample only.

\textbf{Multi-row Processing.}
For simplicity, above we describe the forward and reverse process in terms of a \textit{single row} $z_t$ of the action-class AD image.
However, the same diffusion process can be simultaneously applied to all the rows of the action-class AD image $x^a$, where the forward process adds noise to all rows at once (to generate $\{x^a_1, ..., x^a_T \}$ from the ground truth $x^a_0$), and the reverse process aims to reverse the addition of noise in all rows (i.e., produce $x^a_{t-1}$ from $x^a_t$ at the $t$-th step).
We further note that this Discrete Action-Detection Diffusion Process is also used for the starting point and ending point AD images, since they all tackle a frame-wise classification task.

\subsection{Row-Column Transformer Architecture}
\label{sec:transformer_architecture}

Moreover, different from natural images which contain rich local 2D spatial relationships, our AD images possess different relationship patterns along each of the two dimensions (i.e., between frames vs. between classes). 
Specifically, there is a strong natural sequential ordering between adjacent rows (i.e., between adjacent temporal frames), which yet differs from the inter-class relationships between adjacent columns (e.g., between adjacent action classes).
Hence, existing diffusion network designs \citep{ho2020denoising,song2021denoising}, which tend to focus on 2D spatial processing in neighbourhoods, are not well-suited for our use.
To overcome this, we propose to handle the two dimensions in different ways, such that our Row-Column Transformer design can effectively extract class information across the columns while encoding temporal information across the rows.
Below, for simplicity, we describe the architecture of our diffusion network $d$ to handle a single AD image, using the action-class AD image as an example.

Next, we describe the inputs to the diffusion network $d$ at the $t$-th diffusion step.
We denote the input image as $x^a \in  \mathbb{R}^{N \times C}$.
In order to derive predictions specific to the input video, we also extract spatio-temporal features $f_{ST}  \in \mathbb{R}^{N \times C_{ST}}$ from the input video.
By conditioning on $f_{ST}$, our reverse process receives important video-specific information to perform the denoising.
Besides, to better capture the distribution characteristics at each $t$-th step, we also condition the diffusion process on step index $t$, and generate a diffusion step embedding $f_t \in \mathbb{R}^{N \times 1}$ via the sinusoidal function to represent the $t$-th diffusion step.
Then, we concatenate $x^a,f_{ST}$ and $f_t$ to form input $x \in \mathbb{R}^{N \times (C + C_{ST} + 1) }$.

Our Row-Column Transformer design for our diffusion network $d$ consists of $L$ stacks of \textit{Row-Column Blocks}, which we introduce below.
Refer to Supp for more details.

\textbf{Row-Column Block.}
The first part of the block encodes information across columns, i.e., class information.
Crucially, relationships between the columns (i.e., inter-class relationships) can exist over long ranges.
Hence, in order to encode the relationships between columns (classes) across a long range, we perform a Multi-Head Self-Attention (MHSA) between the columns of the input image.
Note that, for starting and ending point AD images, the two neighbouring columns (yes/no) are highly correlated, and these correlations can still be learned with the diffusion design in the previous section and the MHSA operation.
Specifically, we treat each column of the input $x \in \mathbb{R}^{N \times (C+C_{ST}+1) }$ as a token, thus obtaining $C+C_{ST}+1$ tokens of length $N$.
Next, a learnable positional embedding is added to each token, which encodes the positional information of each token.
Then, we perform MHSA among all the $C + C_{ST} +1$ tokens, to obtain an intermediate output $u_{col} \in \mathbb{R}^{N \times (C+ C_{ST}+1) }$.
This is followed by 2 MLP layers, where we eventually output $x_{col} \in \mathbb{R}^{N \times (C+ C_{ST} + 1) }$.

In the next part of the Row-Column Block, we encode temporal information across rows (frames).
Notably, there exist \textit{strong local relationships and sequential correlations} between neighbouring rows (frames).
Furthermore, actions often provide context information for other actions in the same sequence, which we can exploit by considering the \textit{longer-term temporal relationships} between rows (frames).
Thus, to effectively encode local relationships between neighbouring rows (frames), we apply a Temporal Convolution (TC), which has a strong inductive bias for encoding local sequential relationships \citep{dai2022ms,d2021convit}.
We also combine the TC with a MHSA conducted between the rows (frames) to encode long-range temporal relationships.
Specifically, we first process the local relationships with a $1 \times 3$ TC, to yield $u_{row}  \in \mathbb{R}^{N \times (C+ C_{ST}+1)}$.
Then, to encode long-range temporal relationships, we first add a learnable positional embedding to each token, before performing MHSA across the rows (frames) by treating each row of $u_{row}$ as a token (i.e., there are $N$ tokens of length $C+ C_{ST}+1$).
This is followed by 2 MLP layers to obtain a final output $x_{row} \in \mathbb{R}^{N \times (C+ C_{ST}+1) }$.

\textbf{Combined Image Processing.}
Above, we describe both the Discrete Action-Detection Diffusion Process and Row-Column Transformer for the action-class AD image $x^a$.
Yet, these methods can also handle other AD images, since the three AD images (action-class $x^a$, starting point $x^s$, ending point $x^e$) are all similarly designed to perform classification.

To produce the three AD images, one possible way is to perform our methods once for each AD image, i.e., producing them separately.
Another option is to stitch the images together into a \textit{combined image} and perform a combined processing for all three AD images.
Specifically, we can concatenate the three AD images $\{x_t^a,x_t^s,x_t^e \}$ horizontally to obtain the combined image as $x^{combined}_t \in  \mathbb{R}^{N \times(C+4) }$, where each row consists of three discrete distributions.
At the same time, we can still handle the three discrete distributions separately during the diffusion process. 
To process the stitched image with the diffusion network, we only need to modify the column dimensionality, i.e., by adding 4 columns to the Row-Column Block design above.

There are two advantages in such combined processing. Firstly, it is more efficient and allows us to produce the three AD images in parallel at one go.
Secondly, the combined processing also facilitates more sharing of knowledge between the classification and starting/ending point localization sub-tasks, that can be learned via the combined end-to-end update.
For instance, by producing the three AD images simultaneously, our model learns to tackle action classification with the knowledge of the global temporal structure of all action instances in the video (gained from the localization sub-tasks), which leads to better performance.

\subsection{Inference and Training Pipeline}

\textbf{Inference Pipeline.}
After obtaining the outputs $\hat{x}^a_0$, $\hat{x}^s_0$, $\hat{x}^e_0$, we use them to obtain the final action instances.
We largely follow the post-processing pipeline of \citep{lin2018bsn,lin2019bmn}, as follows:
First, we find the frames where actions are likely to start or end, by finding pixels in the left column of $\hat{x}^s_0,\hat{x}^e_0$ that are above a pre-defined threshold $\delta$. 
Next, because these pixels are often found in clusters, we group up the pixels that are connected, and identify their average position as the starting or ending point. 
Then, following previous approaches \citep{lin2018bsn,lin2019bmn} to generate candidate proposals, each starting point $\hat{x}^s_0$ is coupled with all the ending points behind to identify action candidates,
with the action duration ranging between the identified starting and ending points. To obtain the action candidate's class, we average the rows of $\hat{x}^a_0$ corresponding to the duration of the action candidate, and take the class with the highest value. After obtaining all the action candidates for the video sequence, we utilize Soft-NMS \citep{bodla2017soft} to remove overlapping candidates and produce the final results.

\textbf{Training Pipeline.} 
We use an off-the-shelf model to extract video features $f_{ST}$, which is kept frozen throughout.
At the start, we randomly initialize the diffusion model $d$ with the architecture in Sec.~\ref{sec:transformer_architecture}.
During training, we obtain supervision signals for the intermediate steps via the forward process (Eq.~\ref{eq:zt from z0_2}).
Then, we perform the reverse process (Eq.~\ref{eq: sample}) with our diffusion model $d$ to obtain AD image predictions for the intermediate steps and output.
We apply the MSE loss between our AD image predictions and the supervision signals at each step, to update the parameters of the diffusion model $d$.

\section{Experiments}

\subsection{Implementation Details}

Following existing works \cite{zhang2022actionformer,shi2023tridet}, we use an off-the-shelf I3D \cite{carreira2017quo} and R(2+1)D \cite{tran2018closer} models to extract video features $f_{ST}$ for THUMOS14 and ActivityNet-1.3 respectively.
The diffusion network $d_{\phi}$ is randomly initialized following the Xavier initialization scheme \citep{glorot2010understanding}. 
Following previous image diffusion works \cite{ho2020denoising,song2021denoising}, we adopt MSE loss for training. We use AdamW \citep{loshchilov2017decoupled} as the optimizer.
The initial learning rate is set to $2\times10^{-5}$ and decays following cosine rule. 
The training batch size is set to 16. Following previous work \citep{zhang2022actionformer}, all training videos are padded to be 2,304 frames and mask operations are added accordingly for redundant padded frames. 
We set the hyperparameters $T=50$, $L=3$, $\delta=0.9$, $M= 10$.
$K = 200$ for THUMOS14 and $K = 2000$ for ActivityNet-1.3.
All experiments are conducted on Nvidia V100 GPUs. Refer to Supp for more implementation details.

\subsection{Datasets and Evaluation Metric}

Following previous works \citep{shi2023tridet,nag2023difftad,nag2022proposal,zhang2022actionformer,shi2022react,weng2022efficient,lin2019bmn,xu2020g}, we evaluate our method on the THUMOS14 and ActivityNet-1.3 datasets.
\textbf{THUMOS14} \citep{idrees2017thumos} includes 413 untrimmed videos containing 20 classes of actions. The THUMOS14 dataset is split into a validation set with 200 videos and a test set with 213 videos. We follow existing settings \citep{wang2016actionness,zhang2022actionformer,lin2021learning,bai2020boundary} to train our model on the validation set and test on the test set.
\textbf{ActivityNet-1.3} \citep{caba2015activitynet} contains over 20K videos and 200 action categories. 
It comprises of training, validation and test splits containing 10,024, 4,926 and 5,044 videos respectively. 
Following previous settings \citep{xu2020g,lin2019bmn,lin2018bsn,zhang2022actionformer}, our model is optimized on the training set and tested on the validation set.

\textbf{Evaluation Metric.}
Following previous works \citep{zhang2022actionformer,shi2022react,weng2022efficient,nag2022proposal}, we report the mean average precision (mAP) at different temporal intersection over union (tIoU) thresholds.
The tIoU threshold determines how much overlap is required between the prediction and the ground truth to be considered an accurate prediction.
We also report the average mAP (Avg), where we average across several tIoUs.

\begin{table}[t]
\tiny
\centering
\caption{Results on THUMOS14 and ActivityNet-1.3 datasets.
}
\vspace{-3mm}
\tabcolsep=1mm
\begin{tabular}{l|cccccc>{\columncolor[gray]{0.85}}c|cccc>{\columncolor[gray]{0.85}}c}
        \hline
         & \multicolumn{7}{|c|}{THUMOS14} &  \multicolumn{5}{c}{ActivityNet-1.3} \\
        Methods & Feature & 0.3 & 0.4 & 0.5 & 0.6 & 0.7 & Avg & Feature & 0.5 & 0.75 & 0.95 & Avg  \\ 
        \hline
        BMN \citep{lin2019bmn} & TSN & 56.0 & 47.4 & 38.8 & 29.7 & 20.5 & 38.5  & TSN & 50.1 & 34.8 & 8.3 & 33.9  \\
        DBG \citep{lin2020fast} & TSN & 57.8 & 49.4 & 39.8 & 30.2 & 21.7 & 39.8  & - & - & - & - & - \\
        G-TAD \citep{xu2020g} & TSN & 54.5 & 47.6 & 40.3 & 30.8 & 23.4 & 39.3  & TSN & 50.4 & 34.6 & 9.0 & 34.1  \\
        BC-GNN \citep{bai2020boundary} & TSN & 57.1 & 49.1 & 40.4 & 31.2 & 23.1 & 40.2  & TSN & 50.6 & 34.8 & 9.4 & 34.3 \\
        TAL-MR \citep{zhao2020bottom} & I3D & 53.9 & 50.7 & 45.4 & 38.0 & 28.5 & 43.3   & I3D & 43.5 & 33.9 & 9.2 & 30.2 \\
        P-GCN \citep{zeng2019graph} & R(2+1)D & 69.1 & 63.3 & 53.5 & 40.5 & 26.0 & 50.5  & I3D & 48.3 & 33.2 & 3.3 &  31.1 \\
        TSA-Net \citep{gong2020scale} & P3D & 61.2 & 55.9 & 46.9 & 36.1 & 25.2 & 45.1   & P3D & 48.7 & 32.0 & 9.0 & 31.9 \\
        MUSES \citep{liu2021multi} & I3D & 68.9 & 64.0 & 56.9 & 46.3 & 31.0 & -  & I3D & 50.0 & 35.0 & 6.6 & 34.0  \\
        TCANet \citep{qing2021temporal} & TSN & 60.6 & 53.2 & 44.6 & 36.8 & 26.7 & 44.3   & TSN & 52.3 & 36.7 & 6.9 & 35.5  \\
        BMN-CSA \citep{sridhar2021class} & TSN & 64.4 & 58.0 & 49.2 & 38.2 & 27.8 & 47.7  & TSN & 52.4 & 36.2 & 5.2 & 35.4   \\
        ContextLoc \citep{zhu2021enriching} & I3D & 68.3 & 63.8 & 54.3 & 41.8 & 26.2 & 50.9   & I3D & 56.0 & 35.2 & 3.6 & 34.2  \\
        VSGN \citep{zhao2021video} & TSN & 66.7 & 60.4 & 52.4 & 41.0 & 30.4 & 50.2  & R(2+1)D & 53.3 & 36.8 & 8.1 & 35.9  \\
        RTD-Net \citep{tan2021relaxed} & I3D & 68.3 & 62.3 & 51.9 & 38.8 & 23.7 & 49.0   & I3D & 47.2 & 30.7 & 8.6 & 30.8  \\
        A$\textsuperscript{2}$Net \citep{yang2020revisiting} & I3D & 58.6 & 54.1 & 45.5 & 32.5 & 17.2 & 41.6 & I3D & 43.6 & 28.7 & 3.7 & 27.8 
 \\
        GTAN \citep{long2019gaussian} & P3D & 57.8 & 47.2 & 38.8 & - & - & -   & P3D & 52.6 & 34.1 & 8.9 & 34.3  \\
        PBRNet \citep{liu2020progressive} & I3D & 58.5 & 54.6 & 51.3 & 41.8 & 29.5 & -  & I3D & 54.0 & 35.0 & 9.0 & 35.0  \\
        TadTR \citep{liu2022end} & I3D & 62.4 & 57.4 & 49.2 & 37.8 & 26.3 & 46.6   & I3D & 49.1 & 32.6 & 8.5 & 32.3  \\
        AFSD \citep{lin2021learning} & I3D & 67.3 & 62.4& 55.5 & 43.7 & 31.1 & 52.0   & I3D & 52.4 & 35.3 & 6.5 & 34.4  \\
        TAGS \citep{nag2022proposal} & I3D & 68.6 & 63.8 & 57.0 & 46.3 & 31.8 & 52.8  & I3D & 56.3 & 36.8 & 9.6 & 36.5  \\        
        STPT \citep{weng2022efficient} & STPT & 70.6 & 65.7 & 56.4 & 44.6 & 30.5 & 53.6   & STPT & 51.4 & 33.7 & 6.8 & 33.4  \\
        ReAct \citep{shi2022react} & 3DCNN & 69.2 & 65.0 & 57.1 & 47.8 & 35.6 & 55.0   & 3DCNN & 49.6 & 33.0 & 8.6 & 32.6   \\
        ActionFormer \citep{zhang2022actionformer} & I3D & 82.1 & 77.8 & 71.0 & 59.4 & 43.9 & 66.8   & R(2+1)D & 54.7 & 37.8 & 8.4 & 36.6 \\
        DiffTAD \citep{nag2023difftad}  & I3D & 74.9 & 72.8 & 71.2 & 62.9 & 58.5 & 68.0 & I3D & 56.1 & 36.9 & 9.0 & 36.1     \\        
        Self-DETR \citep{kim2023self} &  I3D  &  74.6  &  69.5  & 60.0 & 47.6  &  31.8  & 56.7 & I3D &  52.2 & 33.6 & 8.4 & 33.7  \\
        TriDet \citep{shi2023tridet} & I3D & 83.6 & 80.1 & 72.9 & 62.4 & 47.4 & 69.3  & R(2+1)D & 54.7 & 38.0 & 8.4 & 36.8\\        
        \hline
        Ours  & I3D & 84.9 &  81.5 & 76.5 & 63.0 & 48.0 & \textbf{70.8}  & R(2+1)D & 56.9 & 38.9 & 9.1 & \textbf{38.3}  \\        
        \hline
\end{tabular}
\label{tab:main_results}
\vspace{-2mm}
\end{table}

\subsection{Main Experimental Results}

We compare with state-of-the-art action detection methods on THUMOS14 and ActivityNet-1.3 datasets in Tab.~\ref{tab:main_results}. 
Our proposed method achieves the best results on average mAP among existing methods, showing its efficacy.

\subsection{Ablation Studies}
Following previous works \citep{zhang2022actionformer,weng2022efficient,shi2022react,shi2023tridet},
we conduct ablation experiments on THUMOS14.

\noindent
\textbf{Impact of Main Components of ADI-Diff Framework.}
First, we evaluate the efficacy of our proposed Discrete Action-Detection Diffusion process by comparing against the following baselines: 
\textbf{(A) Stand. Diff. + Model Architecture from \citep{ho2020denoising}}: We apply the standard diffusion process \citep{ho2020denoising} with the image diffusion model architecture from \citep{ho2020denoising}.
\textbf{(B) Disc. AD Diff. + Model Architecture from \cite{ho2020denoising}}: We adopt our Discrete Action-Detection Diffusion, but use the image diffusion model architecture from \citep{ho2020denoising}.
\textbf{(C) Stand. Diff. + Row-Col}: We apply the standard diffusion process \citep{ho2020denoising} with our Row-Column Transformer.
As observed in Tab.~\ref{tab:main_ablation}, Baseline B which uses the proposed diffusion process obtains a much better result than Baseline A which uses standard diffusion, showing the efficacy of the proposed diffusion process. Notably, this trend also persists when the Row-Column Transformer is used, where our method significantly outperforms Baseline C.
This improvement is because our Discrete Action-Detection Diffusion allows us to effectively map the noisy distributions to the underlying target distribution.

Next, we also validate the efficacy of our proposed Row-Column Transformer design.
First, we compare Baseline A vs Baseline C, as well as Baseline B vs our method, and find that the proposed Row-Column Transformer leads to performance improvements in both cases, no matter if the standard diffusion or our proposed diffusion process is used.
This shows the efficacy of the proposed Row-Column Transformer.
Besides, we further ablate the design of the Row-Column Transformer by comparing against the following alternative designs while applying our proposed diffusion process:
\textbf{(D) Disc. AD Diff. + Model Architecture from \citep{rombach2022high}} adopts the model architecture from \citep{rombach2022high};
\textbf{(E) Disc. AD Diff. + Row-Col (w/ Col design only)} adopts an alternative network design (with approximately same network size) that processes both the columns and rows the same way, with MHSA layers only;
\textbf{(F) Disc. AD Diff. + Row-Col (w/ Row design for both)} adopts an alternative network design (with approximately same network size) that processes both the columns and rows the same way, with TC+MHSA layers;
\textbf{(G) Disc. AD Diff. + Row-Col (w/o Learnable PE)} replaces the learnable positional embedding with a fixed one generated using the sinusoidal function.
Overall, as shown in Tab.~\ref{tab:main_ablation}, our proposed design performs the best, showing its efficacy in capturing both class-wise (across columns) and temporal relationships (across rows).

\begin{table}[h]
    \scriptsize
    \centering
    \caption{
    Ablation study for main components of ADI-Diff.  
    }    
    \vspace{-3mm} 
    \tabcolsep=0.95mm
    \begin{tabular}{l|c|c|c|c}
        \hline
        Method & 0.3  & 0.5 & 0.7 & Avg \\
        \hline
        (A) Stand. Diff. + Model Architecture from \cite{ho2020denoising} & 80.4  & 68.2  & 44.1 & 66.0   \\
        (B) Disc. AD Diff. + Model Architecture from \cite{ho2020denoising}  & 82.6  & 74.3 & 45.2 & 69.0 \\        
        (C) Stand. Diff. + Row-Col  & 82.1 & 73.9 & 45.0 & 68.1 \\
        \hline
        (D) Disc. AD Diff. + Model Architecture from \citep{rombach2022high} & 82.0 & 74.1 & 45.2 & 67.5 \\        
        (E) Disc. AD Diff. + Row-Col (w/ Col design for both) & 81.8  & 75.5  & 45.0 & 69.0 \\ 
        (F) Disc. AD Diff. + Row-Col (w/ Row design for both) & 82.0  & 75.3  & 45.1 & 68.8 \\       
        (G) Disc. AD Diff. + Row-Col (w/o Learnable PE) & 82.4  & 75.2 & 45.9 & 69.1 \\ 
        \hline
        Ours (Disc. AD Diff. + Row-Col) & 84.9  & 76.5  & 48.0 & 70.8 \\
        \hline
    \end{tabular}
    \label{tab:main_ablation}
\end{table}

\begin{figure}[h]
    \centering
    \vspace{1mm}
    \includegraphics[width=0.85\linewidth]{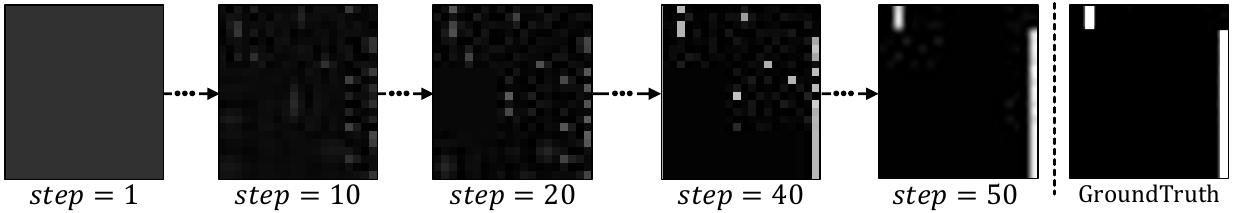}
    \vspace{-0.3cm}
    \caption{
    Visualization of diffusion process.
    }
    \label{fig:reverse}
\end{figure}

\noindent
\textbf{Visualization of Diffusion Process.}
In Fig.~\ref{fig:reverse}, we visualize the action-class AD images produced throughout the reverse diffusion process. 
We observe that our ADI-Diff framework progressively denoises the original noisy discrete probability distributions, to produce high-quality discrete action-class distributions.
See Supp for more results.

\begin{wrapfigure}{r}{0.38\linewidth}
    \centering
    \includegraphics[width=\linewidth]{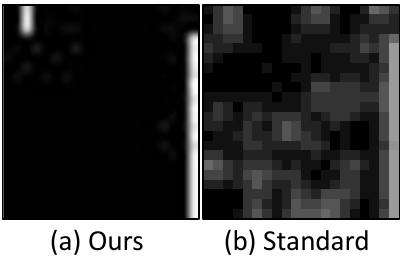}
    \vspace{-0.8cm}
    \caption{
    Comparison between the action-class AD image generated by our method (left) and standard diffusion (right).
    }
    \label{fig:gau_uni_vis}
\end{wrapfigure}

We also compare against a \textbf{Standard} baseline, which applies a standard diffusion process \citep{ho2020denoising}.
As observed in Fig.~\ref{fig:gau_uni_vis}, our proposed method produces better predictions qualitatively as compared to the Standard baseline.
Specifically, the baseline's action-class AD image (right of Fig.~\ref{fig:gau_uni_vis}) shows much ambiguity and confusion, where the class predictions can be spread over multiple columns (i.e., white pixels 
\setlength{\columnsep}{4pt}%
are not concentrated in a consistent column).
In contrast, the action-class AD image produced by our method (left of Fig.~\ref{fig:gau_uni_vis}) tends to consistently provide the correct action-class -- here, the predictions are concentrated on two separate columns, indicating that two action-classes are captured in this video clip, with one action happening after the other.

\noindent
\textbf{Impact of Image Stitching.} 
Next, we explore the impact of stitching three of our AD images into a combined image for processing. Results are shown in Tab.~\ref{tab:ablation_stitching}. 
We find that stitching the images leads to an efficiency gain and some accuracy gains, as it allows us to produce the AD images in parallel while also facilitating the sharing of knowledge.

\begin{table}[h]
    \scriptsize
    \centering
    \caption{Ablation study for stitching AD images.
    }
    \vspace{-0.3cm}    
    \begin{tabular}{c|c|c|c|c|c}
    \hline  
    Setting & 0.3 & 0.5 & 0.7 & Avg & Speed (seconds per clip) \\
    \hline
    w/o stitching & 84.7 & 76.0 & 47.8 & 70.3 & 0.158 \\      
    \hline
    w/ stitching & 84.9 & 76.5 & 48.0 & 70.8 & 0.113 \\      
    \hline
    \end{tabular}
    \label{tab:ablation_stitching}
    \vspace{2mm}
\end{table}

\noindent
\textbf{Impact of Temporal Boundary AD Images $x^s,x^e$.}
We also investigate the impact of introducing $x^s,x^e$ by evaluating the performance without them, where here we follow previous approaches \cite{zhang2022actionformer,shi2023tridet} to directly regress the starting and ending points.
In Tab.~\ref{tab:ablation_temporalboundary}, we observe that the performance is much better when we add $x^s,x^e$, which shows the importance of introducing the temporal boundary AD images to handle the action detection task.

\begin{table}[h]
    \scriptsize
    \centering
    \caption{Ablation study for temporal boundary AD images.
    }    
    \vspace{-0.3cm}    
    \begin{tabular}{c|c|c|c|c}
    \hline  
    Setting & 0.3  & 0.5  & 0.7 &  Avg  \\
    \hline
    w/o temporal boundary AD images & 80.8 & 71.5 & 43.9 & 67.3  \\  
    \hline
    w/ temporal boundary AD images & 84.9  & 76.5  & 48.0 & 70.8 \\ 
    \hline
    \end{tabular}
    \label{tab:ablation_temporalboundary}
    \vspace{2mm}
\end{table}

\noindent
\textbf{Inference Speed.}
In Tab.~\ref{tab:ablation_inference_speed}, we compare our method's speed against existing methods in terms of seconds per video clip. 
Our method achieves comparable speed to the state-of-the-art \citep{shi2023tridet}, yet significantly outperforms it.

\begin{table}[h]
    \scriptsize
    \centering
    \vspace{1mm}
    \caption{Inference speed.
    }    
    \vspace{-3mm}       
    \begin{tabular}{l|c|c|c|c|c}
        \hline
        Setting & 0.3 & 0.5 &  0.7 & Avg & Speed (seconds per clip) \\        
        \hline
        DiffTAD \citep{nag2023difftad} & 74.9 & 71.2  & 58.5 & 68.0 & 0.397 \\         
        TriDet \citep{shi2023tridet} & 83.6 & 72.9 & 47.4 & 69.3  &  0.110 \\        
        \hline
        Ours & 84.9  & 76.5  & 48.0 & 70.8 & 0.113\\
        \hline
    \end{tabular}
    \label{tab:ablation_inference_speed}   
\end{table}

\section{Conclusion}
In this paper, we tackle action detection by casting it as an image generation problem, and propose an AD Image Diffusion (ADI-Diff) framework to generate target AD images via diffusion.
With a Discrete Action-Detection Diffusion Process and a Row-Column Transformer design, we attain state-of-the-art performance on two widely-used datasets.

\noindent
\textbf{Acknowledgements.}
This project is supported by the Ministry of Education, Singapore, under the AcRF Tier 2 Projects (MOE-T2EP20222-0009 and MOE-T2EP20123-0014), National Research Foundation Singapore under its AI Singapore Programme (AISG-100E-2023-121).


\end{document}